\pdfoutput=1

\documentclass[11pt]{article}

\usepackage{ACL2023}
\usepackage{times}
\usepackage{latexsym}
\usepackage{booktabs}
\usepackage{multirow}
\usepackage{multicol}
\usepackage{xspace}
\usepackage{graphicx}
\usepackage{subcaption}
\usepackage{amsmath}
\usepackage{enumitem}
\usepackage{array}

\usepackage[T1]{fontenc}

\usepackage[utf8]{inputenc}

\usepackage{microtype}

\usepackage{inconsolata}
\usepackage{graphicx}
\usepackage{comment}
\usepackage{algorithm}
\usepackage{algpseudocode}

\usepackage{amssymb, xcolor} 
\newcommand{\greencheck}{{\color{green}\checkmark}} 
\usepackage{pifont}   
\newcommand{\redcross}{{\color{red}\ding{55}}}  

\newenvironment{itemize*}%
 {\leftmargini=12pt\begin{itemize}%
  \setlength{\itemsep}{0pt}%
  \setlength{\parskip}{0pt}%
  }%
 {\end{itemize}}
\newenvironment{enumerate*}%
 {\begin{enumerate}%
  \setlength{\itemsep}{0pt}%
  \setlength{\parskip}{0pt}}%
 {\end{enumerate}}

\newcommand{\modelname}{\textsc{ScaleEval}\xspace}

\title{Can Large Language Models be Trusted for Evaluation? \\ Scalable Meta-Evaluation of LLMs as Evaluators via Agent Debate}

\author{
Steffi Chern\textsuperscript{\rm{2,4}} \quad Ethan Chern\textsuperscript{\rm{1,4}} \quad {Graham Neubig}\textsuperscript{\rm{2}} \quad \textbf{Pengfei Liu}\textsuperscript{\rm{1,3,4}}\thanks{\ \ Corresponding author}\\
\textsuperscript{1}Shanghai Jiao Tong University \
\textsuperscript{2}Carnegie Mellon University \\ 
\textsuperscript{3}Shanghai Artificial Intelligence Laboratory \
\textsuperscript{4}Generative AI Research Lab (GAIR) \
}

\begin{document}
\maketitle

\begin{abstract}
Despite the utility of Large Language Models (LLMs) across a wide range of tasks and scenarios, developing a method for reliably evaluating LLMs across varied contexts continues to be challenging. Modern evaluation approaches often use LLMs to assess responses generated by LLMs. However, the meta-evaluation conducted to assess the effectiveness of these LLMs as evaluators is typically constrained by the coverage of existing benchmarks or requires extensive human annotation. This underscores the urgency of methods for \textit{scalable} meta-evaluation that can effectively, reliably, and efficiently evaluate the performance of LLMs as evaluators across diverse tasks and scenarios, particularly in potentially new, user-defined scenarios. To fill this gap, we propose \modelname, an \textit{agent-debate-assisted meta-evaluation framework} that leverages the capabilities of multiple communicative LLM agents. This framework supports multi-round discussions to assist human annotators in discerning the most capable LLMs as evaluators, which significantly eases their workload in cases that used to require large-scale annotations during meta-evaluation. We release the code for our framework, which is publicly available at: 
\url{https://github.com/GAIR-NLP/scaleeval}.
\end{abstract}

\section{Introduction}
\begin{figure}[!htb]
  \centering
  \includegraphics[width=\linewidth]{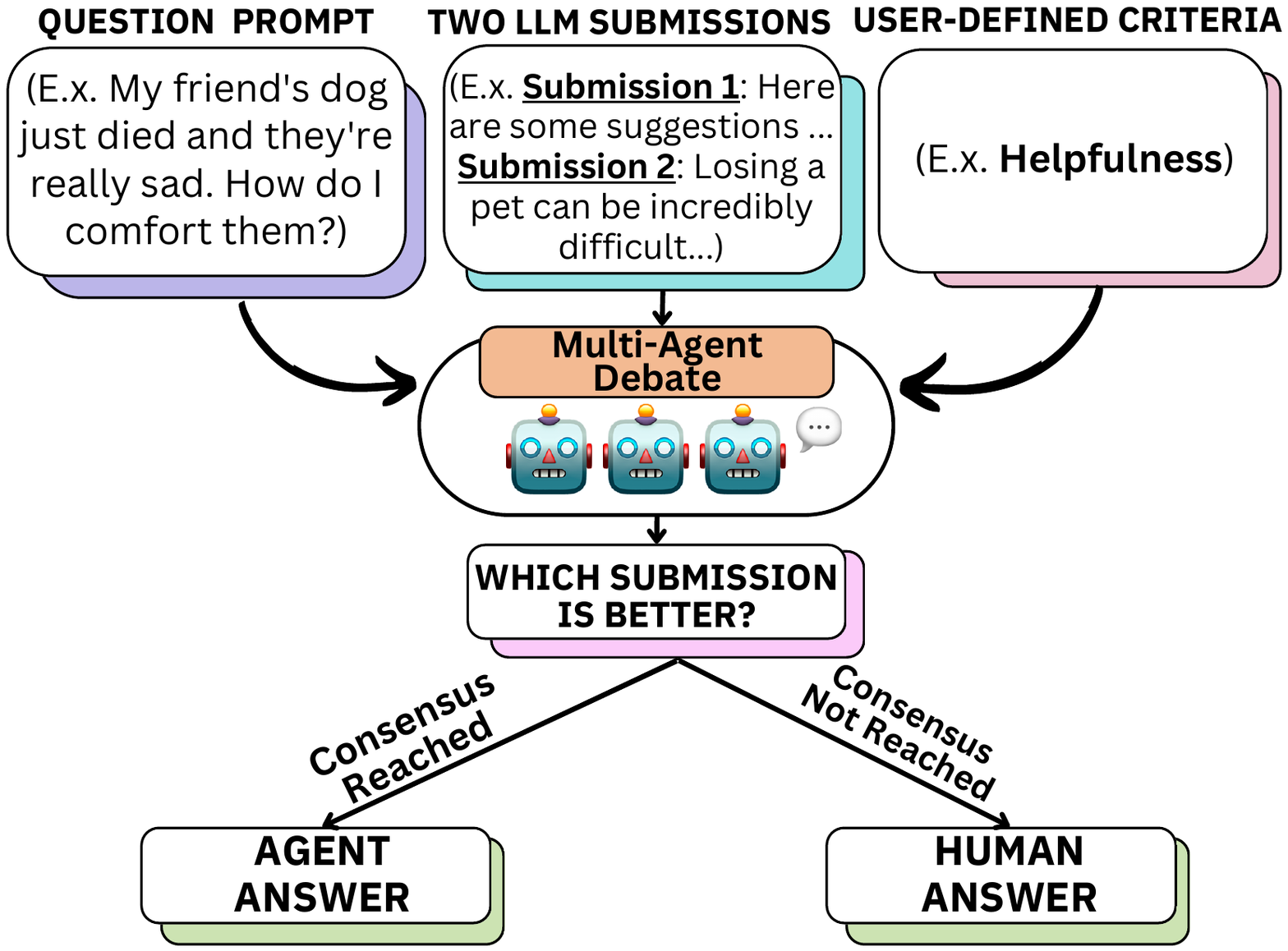}
  \caption{We demonstrate \modelname, our scalable meta-evaluation framework. This is used in assessing the reliability and robustness of employing LLMs as evaluators for different evaluative purposes.} 
  \label{fig:meta_framework}
\end{figure}

Large Language Models (LLMs)~\cite{bubeck2023sparks, team2023gemini} have rapidly evolved to the point where they can tackle a wide range of tasks with impressive performance. While this has unlocked a variety of exciting potential applications, it has also introduced complex challenges in evaluating the generated outputs.
Current efforts on LLM evaluation primarily focus on automated evaluation metrics \citep{fu2023gptscore, alpaca_eval, zheng2023judging, wang2023chatgpt}, many of which use LLMs themselves to do evaluation.
However, when these LLMs as evaluators are applied to a new task, it begs the question: \emph{can LLMs be trusted for evaluation?} In many cases, the answer is not clear.

On the other hand, there are a few fortunate tasks where meta-evaluation (evaluation of evaluation metrics) has been performed rigorously (\S\ref{sec:related_work}).
This meta-evaluation typically involves the collection of human-annotated judgements for particular criteria (e.g.~fluency of outputs, semantic adherence to the input). For instance, for machine translation quality metrics, there is an extensive meta-evaluation data from the WMT metrics task \cite{freitag2022results}, and for summarization there are datasets like TAC and RealSum~\cite{dang2008overview, bhandari2020re}.
Once such a dataset is collected, meta-evaluation can be performed by measuring the correlation between automatic evaluation metrics and the human gold-standard (\S\ref{sec:preliminaries}).

However, these datasets are extremely costly to collect, as they require meticulous annotation by skilled human experts. With the increasing use of LLMs for various purposes such as math problem solving \cite{hendrycks2021measuring}, reading comprehension \cite{AGIEVAL}, creative writing \cite{zheng2023judging}, multilingual applications \cite{XTREME, MULTI}, and many more, it is not feasible to create these human-judged datasets for every new task. As a result, LLMs as evaluators are used without proper vetting, and in many cases the evaluators themselves are highly unreliable \citep{Wang2023LargeLM,huang2023large}.

In this paper, we propose \modelname, a \emph{scalable meta-evaluation framework} for the era of LLMs, which creates meta-evaluation benchmarks across various tasks and scenarios (\S\ref{sec:methodology}).
Concretely, \modelname relies on debate between multiple LLM agents, followed by minimal human oversight in cases where the agent LLMs do not agree (Fig. \ref{fig:meta_framework}). Since our framework allows users to use their own prompts and responses while applying the framework to any scenario or criterion that they define, it offers flexibility and adaptability in various evaluation contexts.

In experiments, we conduct meta-meta evaluation (\S\ref{sec:metametaeval}) demonstrating that our proposed approach correlates well with when meta-evaluation is performed entirely by human expert annotators. Further, we assess the reliability and cost-performance trade-off of various LLMs as evaluators under a variety of scenarios, and closely examine their specific capabilities and limitations as evaluators (\S\ref{sec:metaeval}). We also examine the impact that variations in prompts used for evaluation can have on the performance of LLMs as evaluators (\S\ref{sec:criteria_prompts}).

All code from our framework is made available open-source, enabling the community to conduct meta-evaluation on LLMs as evaluators using their own prompts, LLM responses, criteria, and scenarios.

\section{Related Work}
\label{sec:related_work}
\setlength{\tabcolsep}{2pt}
\begin{table}[t]
  \centering
  \resizebox{\columnwidth}{!}{
    \begin{tabular}{lcccc}
    \toprule
          & \textbf{Meta-Eval} & \textbf{\# Scenarios} & \multicolumn{1}{l}{\textbf{Custom.}} & \textbf{Scala.} \\
    \midrule
    \textbf{LLM-as-a-Judge} & Human & High  &    \redcross   & Low \\
    \textbf{FairEval} & Human & Low   &  \redcross     & Low \\
    \textbf{ChatEval} & Human & Low   &   \redcross    & Low \\
    \textbf{\modelname} & Agent Debate & High  &   \greencheck    & High \\
    \bottomrule
    \end{tabular}%
    }
\caption{Comparison of the meta-evaluation processes across different strategies using LLMs as evaluators: LLM-as-a-Judge \cite{zheng2023judging}, FairEval \cite{Wang2023LargeLM}, ChatEval \cite{chan2023chateval}, and our own work, \modelname. ``Custom.'' denotes whether the evaluation criterion could be customized. ``Scala.'' refers to scalability.}
\label{tab:comparison}
\end{table}%

\subsection{Automatic Evaluation of LLM Output}
The most common paradigm for evaluating LLMs is to evaluate their capabilities on standard benchmarks for tasks such as reasoning (e.g.~BigBench \cite{srivastava2022beyond}), common sense QA (e.g.~MMLU \cite{hendrycks2020measuring}), or code generation (e.g.~HumanEval \cite{chen2021evaluating}).
These are indicative of the capabilities of the models, but do not measure model abilities for open-ended tasks requiring generation of free-form text.

To adapt to the rapid growth in the capabilities of LLMs for open-ended tasks, LLM evaluation has started to shift towards evaluating generated text directly, often using LLMs themselves as evaluators \cite{fu2023gptscore, alpaca_eval, zheng2023judging, wang2023chatgpt}.
In addition, there are a few recent works that perform LLM-based multi-agent debate to improve the fidelity of evaluation \cite{chan2023chateval, li2023prd}.
While these methods take advantage of the instruction-following capabilities and versatility of LLMs, 
directly using LLMs as evaluators or communicative agents out-of-the-box in diverse, unseen user-defined scenarios provides no guarantees with respect to the accuracy of these methods.
We aim to address this issue by introducing scalable meta-evaluation to ensure the reliability of the evaluation protocol under diverse scenarios.

Another widely used evaluation platform, Chatbot Arena \cite{zheng2023judging} supports a crowd-sourcing method to collect diverse user prompts from various scenarios.
However, the process of evaluating LLMs' performance in Chatbot Arena relies heavily on human evaluations, which may not be readily accessible to everyone interested in assessing LLMs' abilities for a specific tasks or scenario.
In addition, the human evaluators involved are not subject to a uniform set of standards or explicit evaluation guidelines, which could lead to biased or imprecise evaluation assessments.

\subsection{Meta-Evaluation of LLMs as Evaluators}
Previous research proposing methods for LLMs as evaluators usually involves conducting meta-evaluation in 3 different ways: (i) leveraging existing NLP meta-evaluation benchmarks \cite{fu2023gptscore, chan2023chateval}, (ii) conducting small-scale meta-evaluations on expert-annotated datasets for specific tasks or scenarios \cite{chiang2023can, wang2023chatgpt, zheng2023judging}, or (iii) using crowd-sourcing platforms to collect human annotations \cite{zheng2023judging}. However, due to the lack of coverage in existing datasets and annotation budgets, both (i) and (ii) are inherently limited in their comprehensiveness. (iii) can provide more comprehensive meta-evaluation via crowd-sourcing, but the amount of human annotation required in the meta-evaluation process limits the scalability of the approach, and crowd workers may not be particularly accurate at more complex tasks.
To address these issues, we propose an agent-debate-assisted meta-evaluation approach to mitigate this effort.

\section{Preliminaries}
\label{sec:preliminaries}

In this section, we provide an introduction to the concepts of automatic evaluation and meta-evaluation systems, particularly focused on evaluation of LLM-generated outputs in the era of generative AI. 

\subsection{Key Terms}
We first define some key terms that will be used throughout our paper. 

\begin{itemize*}
    \item \textbf{Criterion:} A criterion defines a standard that measures the quality of the response generated by LLMs based on the user prompt. Some examples include: helpfulness, fluency, factuality, or creativity, among others. 
    \item \textbf{Scenario:} A scenario describes the real-world situations in which users are interacting with LLMs. For example, brainstorming, coding, and dialog, among others.
\end{itemize*}

\subsection{Automatic Evaluation}
\label{subsec:preliminaries-eval}
Automatic evaluation using LLMs measures the quality of LLM-generated responses given prompts under different criteria. Usually, automatic evaluation is conducted with one of two different protocols: single-response evaluation and pairwise response comparison \cite{ouyang2022training, zheng2023judging, li2023generative}. In this paper, we focus on \textbf{pairwise response comparison}. Pairwise response comparison is intuitive for both humans and LLMs as evaluators when conducting assessments. It could be further extended to provide win-rates and Elo scores across models \cite{zheng2023judging}, offering a straightforward leaderboard to understand the relative performance of different models under various scenarios. Formally, given an automatic evaluation metric $E$, a user-defined evaluation criterion $c$ (e.g. helpfulness, reasoning, creativity), a user prompt $p$, and responses generated by two systems $r_1, r_2$, evaluation for pairwise response comparison is done in the following way:

\begin{equation}
    o = E(c, p, r_1, r_2).
\end{equation}
$o \in \{1, 0, -1\}$ represents that $r_1$ is better, equal, or worse than $r_2$,  respectively, given the user prompt $p$ under criterion $c$.

\subsection{Meta-Evaluation}
\label{subsec:preliminaries-metaeval}
Meta-evaluation assesses the quality of an automatic evaluation metric.
Formally, we define a gold-standard evaluation metric $G$ (e.g. human experts) that other automatic metrics should aspire to match.
In pairwise response comparison, the meta-evaluation dataset $\mathcal{G} = \{ G(c, p_i, r_{1,i}, r_{2,i})\}_{i = 1}^{n}$ contains user prompts and corresponding responses from two systems, annotated with gold-standard evaluations.
The meta-evaluation process assesses the performance $\textsc{meta}(E)$ of the automatic evaluation metric $E$ under a certain criterion $c$.

In pairwise response comparison, the meta-evaluation measures the \textit{example-level agreement rate} or the \textit{system-level agreement rate} between $E$ and $G$ across the meta-evaluation dataset. A high agreement rate between $E$ and $G$ represents that $E$ is a good automatic evaluation metric. 

For the \textit{example-level agreement rate}, we calculate: 
\begin{equation}
    \textsc{meta}(E) =
    \frac{1}{n} \sum_{i = 1}^n \delta_{E(c, p_i, r_{1,i}, r_{2,i}), G(c, p_i, r_{1,i}, r_{2,i})},
\end{equation}
where $0 \leq \textsc{meta}(E) \leq 1$, and $\delta_{\cdot, \cdot}$ refers to the Kronecker delta function. 

For the \textit{system-level agreement rate}, given that $\mathcal{E} = \{ E(c, p_i, r_{1,i}, r_{2,i})\}_{i = 1}^{n}$ and $\mathcal{G} = \{ G(c, p_i, r_{1,i}, r_{2,i})\}_{i = 1}^{n}$, we calculate: 
\begin{equation}
    \textsc{meta}(E) =
    \delta_{\mathrm{mode}(\mathcal{E}), \mathrm{mode}(\mathcal{G})},
\end{equation}
where $\textsc{meta}(E) \in \{0, 1\}$, $\delta_{\cdot, \cdot}$ refers to the Kronecker delta function, and $\mathrm{mode(\cdot)}$ refers to the value (either $1, 0, -1$ in this case) that appears most often in the set $\mathcal{\mathcal{E}}$ or $\mathcal{\mathcal{G}}$.

\section{Methodology}
\label{sec:methodology}
In this section, we detail the frameworks that \modelname employs for meta-evaluation, evaluation, and human expert meta-meta evaluation. For meta-evaluation, we generally follow the pairwise response comparison setting described in \S\ref{subsec:preliminaries-metaeval}. Notably, instead of relying solely on human labor to construct the meta-evaluation benchmark $\mathcal{G}$, we use a scalable, agent-debate assisted framework to instantiate the golden metric $G$ and construct the benchmark $\mathcal{G}$. For evaluation, we follow the pairwise response comparison setting outlined in \S\ref{subsec:preliminaries-eval}. The meta-meta evaluation process also follows the rules for meta-evaluation, as described in \S\ref{subsec:preliminaries-metaeval}. The process is included to ensure the reliability of using the agent-debate assisted meta-evaluation framework.

\subsection{Meta-Evaluation Framework via Multi-Agent Debate}
\label{subsec:methodology-metaeval}
The meta-evaluation framework involves multiple communicative agents $\{A_j\}_{j = 1}^m$ that conduct rounds of discussion $d = 0 \sim D - 1$ with each other. This is less time-consuming and costly compared to traditional methods for meta-evaluation that relies entirely on human effort. With this agent-debate-assisted meta-evaluation framework, we can leverage each LLM agent's distinct understanding about each query prompt $p_i$, LLM responses $r_{1,i}, r_{2,i}$, and defined criterion $c$ to make a comprehensive assessment of LLMs under different scenarios and criteria. Each LLM agent is capable of providing an evaluation result regarding which response is better, along with its corresponding justifications. Note that each LLM agent can also review other agents' evaluation results and justifications after the initial round of discussion.

In the initial round of discussion $d = 0$, each LLM agent independently provides an evaluation result and justification:
\begin{multline}
\mathcal{A}_0 = [A_1(c, p_i, r_{1,i}, r_{2,i}, \varnothing), \ldots, \\ A_m(c, p_i, r_{1,i}, r_{2,i}, \varnothing)],
\end{multline}
where 
\begin{equation}
\mathcal{A}_0[j]_{j = 1, \ldots, m} \in (\{1, 0, -1\}, \textsc{justification}),
\end{equation}
indicates whether $r_{1,i}$ is better, equal, or worse than $r_{2,i}$, respectively, along with its justification. Note that the $\varnothing$ in the last argument of $A_j$ represents that in the initial round of discussion, each agent doesn't have access to previous rounds of discussion. In subsequent discussion rounds $d = 1 \sim D - 1$, agents are allowed to look at other agents' previous assessments and conduct re-evaluations, in which each agent is prompted to stick with or change their original evaluation result. Specifically, given $\mathcal{A}_{d - 1} (d \geq 1)$, which represents the evaluation results and justifications of agents after $(d - 1)^{th}$ rounds of discussions, we conduct the $d^{th}$ round of discussion:
\begin{multline}
\mathcal{A}_d = [A_1(c, p_i, r_{1,i}, r_{2,i}, \mathcal{A}_{d - 1}), \ldots, \\ A_m(c, p_i, r_{1,i}, r_{2,i}, \mathcal{A}_{d - 1})] 
\end{multline}
where similarly to $\mathcal{A}_0$,
\begin{equation}
\mathcal{A}_d[j]_{j = 1, \ldots, m} \in (\{1, 0, -1\}, \textsc{justification}),
\end{equation} The detailed prompt template for meta-evaluation can be found in Table \ref{tab:metaeval-prompt} under Appendix. 

\begin{figure*}[!htbp]
  \centering
  \includegraphics[width=\linewidth, height=11cm]{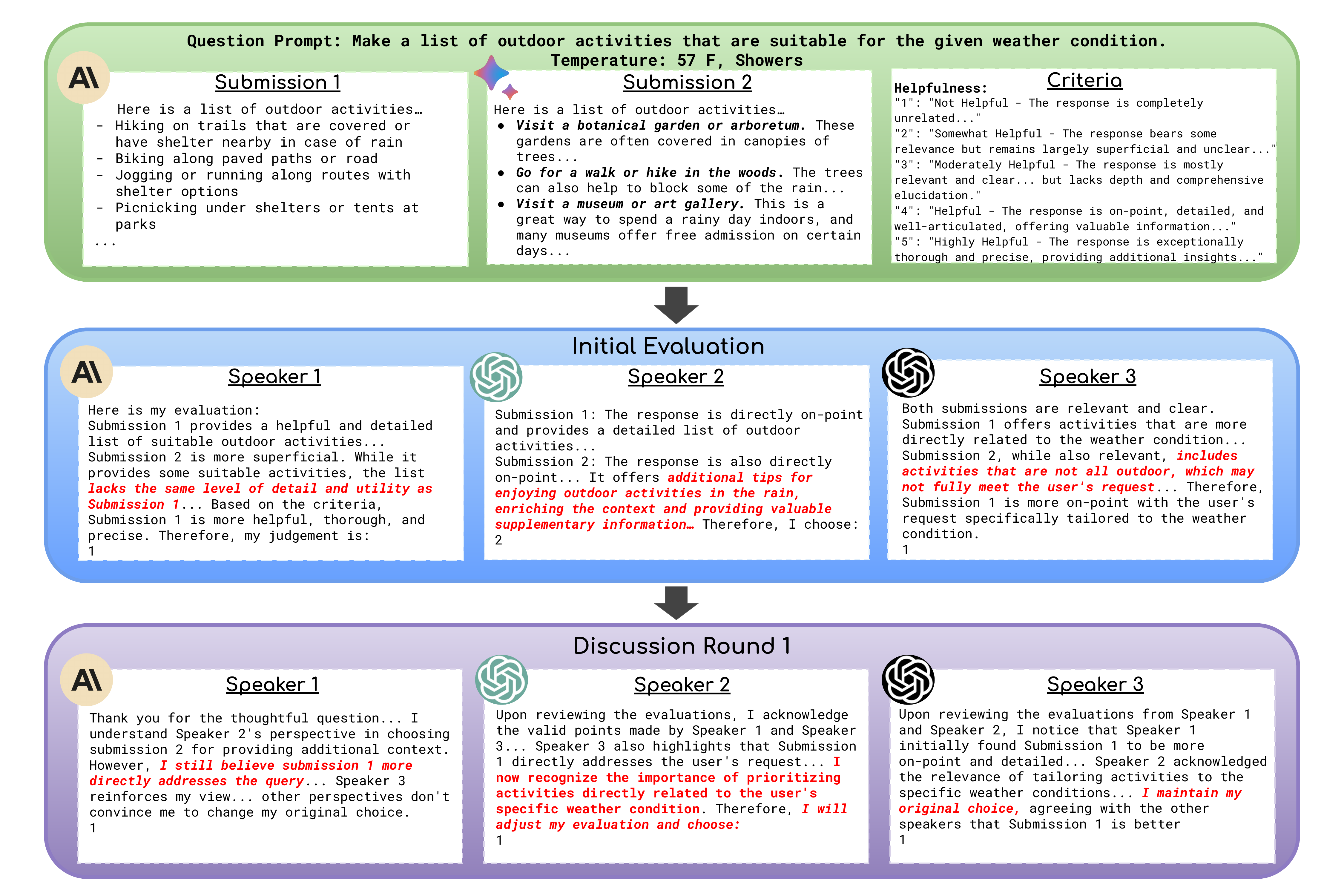}
  \caption{An example of the multi-agent debate process during meta-evaluation.}
  \label{fig:debate}
\end{figure*}

In cases where agents fail to reach a consensus after $d = D - 1$ rounds of discussions, a human evaluator intervenes.
The human evaluator reviews the assessment reports provided by the agents and makes a final decision. Through this process, we incorporate an element of human oversight, thereby increasing the reliability of the final decision. This approach strikes a balance between efficiency and the need for human judgment, ensuring that evaluations are done in a timely and accurate manner. An example of the multi-agent debate process during meta-evaluation is demonstrated in Fig. \ref{fig:debate}.

\subsection{Evaluation Framework}
We follow the pairwise response comparison setting outlined in \S\ref{subsec:preliminaries-eval}. Note that in the LLM era, the automatic evaluation metric $E$ is often instantiated through single LLMs \cite{fu2023gptscore, alpaca_eval, zheng2023judging, wang2023chatgpt}, or multi-agent debate \cite{chan2023chateval, li2023prd}. In \modelname, we focus on instantiating $E$ through single LLMs (e.g., \textit{gpt-3.5-turbo}). However, it is important to note that our framework can be further generalized to other instantiations of $E$.

\subsection{Human Expert Meta-Meta Evaluation}
\label{human expert framework}
To test the reliability of our proposed meta-evaluation framework, we apply meta-meta evaluation. The meta-meta evaluation process also follows the meta-evaluation process described in \S\ref{subsec:preliminaries-metaeval}, where $E$ is instantiated as the agent-debated assisted protocol as described in \S\ref{subsec:methodology-metaeval}, and $G$ is instantiated as the human expert annotation protocol. 

\section{Examined Scenarios}
Establishing real-life scenarios that reflect individuals' daily usage is key to assess the performance and limitations of LLMs in a comprehensive manner. In the current instantiation of \modelname, we include 8 different scenarios that are closely related to everyday situations and tasks \cite{liang2022holistic, li2023generative}. Some example prompts for each defined scenario is shown in Table \ref{tab:scenario_examples}.
We describe more about exactly how we collect data for each of these scenarios below. Individuals interested in evaluating LLMs with our framework can supplement their assessment with additional scenarios.

\begin{table*}[t!]
    \centering
    \small
    \resizebox{\linewidth}{!}{
    \begin{tabular}{c|p{12cm}}
        \toprule
        \textbf{Scenario} & \textbf{Examples} \\
        \midrule
        \multirow{2}{*}{Brainstorming} & \textit{ - Can you tell me how to make chocolate chip cookies?} \\
         & \textit{ - Make a list of snacks and foods to serve as party snacks on a game day!} \\
        \midrule
        \multirow{2}{*}{Coding} & \textit{ - What is the difference between HTML and JavaScript?} \\
         & \textit{ - Implement a binary search algorithm to find a specific element in a sorted array.} \\
        \midrule
        \multirow{2}{*}{Dialog} & \textit{ - Act as the Norse Goddess Freyja.} \\
         & \textit{ - Can you think and feel like a human?} \\
        \midrule
        \multirow{2}{*}{Judgement} & \textit{ - What if the Aztecs had successfully repelled the Spanish conquistadors?} \\
         & \textit{ - How can you determine if a person is genuinely interested in a conversation or simply being polite?} \\
        \midrule
        \multirow{2}{*}{Math} & \textit{ - Given that f(x) = 5$x^3$ - 2$x$ + 3, find the value of f(2).} \\
         & \textit{ - If the endpoints of a line segment are (2, -2) and (10, 4), what is the length of the segment?} \\
        \midrule
        \multirow{2}{*}{ODG} & \textit{ - Is there a meaning for Christmas wreaths?} \\
         & \textit{ - What are some of the best universities for studying robotics?} \\
         \midrule
        \multirow{2}{*}{ODS} & \textit{ - What causes the northern lights?} \\
         & \textit{ - What do the different octane values of gasoline mean?} \\
        \midrule
        \multirow{2}{*}{Writing} & \textit{ - Can you help me write a formal email to a potential business partner proposing a joint venture?} \\
         & \textit{ - Take MLK speech "I had a dream" but turn it into a top 100 rap song.} \\
        \bottomrule
    \end{tabular}}
    \caption{Examined scenarios and corresponding selected examples.}
    \label{tab:scenario_examples}
\end{table*}

\paragraph{Brainstorming} The brainstorming scenario is designed to test the LLMs' ability to engage in problem-solving, creative ideation, and generation of insightful responses, especially in situations that require critical thinking and detailed, step-by-step reasoning.

\paragraph{Coding} The code scenario evaluates LLMs' ability to comprehend, produce, and debug code, as well as answering coding-related questions.

\paragraph{Dialog} The dialog scenario measures LLMs' ability to engage with users in a manner that is intuitive, human-like, and dynamic, testing their proficiency through context-sensitive conversations and role-playing that require maintaining a consistent persona throughout a series of interactions.

\paragraph{Judgement} The judgement scenario assesses LLMs‘ ability to make inferences and formulate opinions, including soliciting insights on diverse situations or emotions, and posing questions that require logical thinking or reasoning.

\paragraph{Math} The math scenario evaluates the LLMs' proficiency in understanding and solving mathematical problems, emphasizing their accuracy in tasks ranging from simple calculations to complex reasoning.

\paragraph{Open-Domain General (ODG)} The ODG scenario measures LLMs' proficiency in applying diverse knowledge and exercising reasoning across a wide array of topics, such as answering questions with definitive answers. 

\paragraph{Open-Domain Science (ODS)} The ODS scenario tests the LLMs' application of scientific knowledge, and gauges their ability to accurately interpret and respond to queries related to scientific disciplines like biology, chemistry, physics, astronomy, and more. 

\paragraph{Writing} The writing scenario evaluates LLMs' ability to summarize, translate, and generate various texts, testing their core language processing and production skills. 

\section{Exp-I: Meta-Meta-Evaluation of Multi-Agent Debate}
\label{sec:metametaeval}
In this section, we first perform meta-meta-evaluation, examining whether the meta-evaluation results of using \modelname match closely to those resulting from meta-evaluation using human evaluators.

\paragraph{Setup}
For our \modelname meta-evaluation framework (as described in \S\ref{subsec:methodology-metaeval}), we deploy three LLM agents to perform multi-agent debate: \textit{gpt-4-turbo, claude-2}, and \textit{gpt-3.5-turbo}.\footnote{Results collected in December 2023. Specific models used are: gpt-4-1106-preview, claude-2, and gpt-3.5-turbo-1106.} In our meta-evaluation experiment, we analyze a total of 160 prompts. This set is comprised 137 prompts from AlpacaEval \cite{alpaca_eval}, 10 coding problem prompts from HumanEval \cite{chen2021codex}, and 13 math problem prompts from GSM-Hard \cite{gao2022pal}. We categorize these prompts into four distinct scenarios: \textit{brainstorming, coding, math,} and \textit{writing}, where each scenario contains 40 prompts.

Each scenario is evaluated based on the following criteria, respectively: \textit{helpfulness, interpretability, reasoning}, and \textit{creativity}. 
We evaluate the generated responses from the following three LLMs: \textit{gpt-3.5-turbo, claude-instant,} and \textit{gemini-pro}. We select the above LLMs to evaluate due to their rather similar performances according to past research and public user feedback, which can help us establish a more nuanced understanding of their performance in various real-world scenarios, and to identify specific contexts where one may outperform the others. 

Our meta-meta evaluation involves having human experts annotate which LLM submission they think is better based on a defined criterion during pairwise comparisons. A total of seven human experts were selected from a pool of Carnegie Mellon University students who have the relevant expertise in answering the queries in each scenario. Different groups of three human experts are responsible for answering the prompts in each scenario, where they are assigned to the scenario that relates to their expertise. Each expert received identical instructions for the task -- they were asked to decide which submission is better based on our defined criteria, and for each comparison, label either \textit{0 (neither submission is better)}, \textit{1 (submission 1 is better)}, or \textit{2 (submission 2 is better)}. The label \textit{2} corresponds to the label \textit{-1} as denoted in section \ref{subsec:preliminaries-eval}. The experts were tasked to conduct 30 comparisons for each of the four different scenarios (\textit{brainstorming, coding, math,} and \textit{writing}), based on their corresponding defined criteria (\textit{helpfulness, interpretability, reasoning,} and \textit{creativity}). This results in a total of 120 final judgements. The question prompts, LLM responses, and criteria utilized for human expert annotations were consistent with those used during our meta-evaluation experiment. All the details were presented in a google sheet that allowed experts to record their answers.

\paragraph{Q1: \textbf{\textit{Can LLM agents with multi-agent debate be used as meta-evaluators in new user-defined scenarios?}}}
To validate the reliability of \modelname's meta-evaluation framework, we perform comparisons between the results from human experts and \modelname's multi-agent debate by two key metrics: the \textit{example-level agreement rate} and the \textit{system-level agreement rate}, as mentioned in \S\ref{subsec:preliminaries-metaeval}. The example-level agreement rate measures the proportion of instances where the multi-agent debate results correspond with the human experts judgements. On the other hand, the system-level agreement rate assesses whether the human experts and multi-agents concur in their overall evaluation of which LLMs produce the best responses for each scenario. A high agreement rate in both metrics would suggest a strong reliability and validity of our meta-evaluation framework, indicating that both human and LLM agents consistently recognize and agree on the quality of responses generated by LLMs.
\paragraph{Results}
From Table \ref{table:baseline exp}, we generally observe a higher example-level agreement rate between human experts and \modelname, compared to the agreement rate between human experts and individual LLM  evaluations. The consistently high agreement rates observed suggest that our meta-evaluation framework aligns well with human expert judgments in these areas, indicating a reliable performance of the collective use of LLMs in meta-evaluating complex scenarios. Across all LLM submission comparisons in our experiment, we observe higher agreement rates in decisions between \modelname outcomes and those of human experts, particularly in coding and math scenarios. This observed trend could be attributed to the inherently objective nature of these subjects, which have relatively clear, definitive answers unlike more subjective areas like creative writing. 

\begin{figure*}[!htbp]
    \centering
    \begin{subfigure}[b]{0.31\textwidth}
        \includegraphics[width=\textwidth]{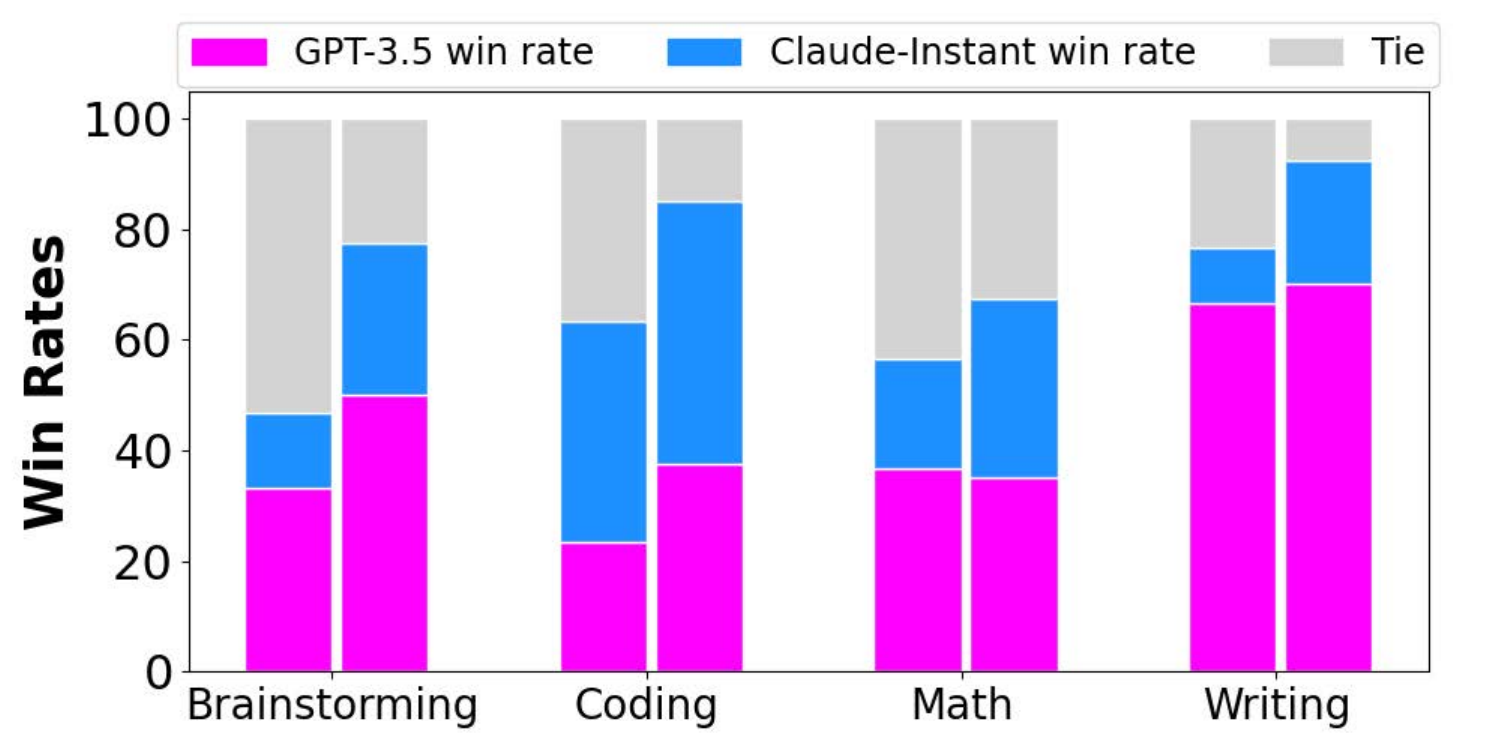}
        \caption{GPT-3.5-Turbo vs. Claude-Instant}
        \label{fig:winrates1}
    \end{subfigure}
    \hfill
    \begin{subfigure}[b]{0.31\textwidth}
        \includegraphics[width=\textwidth]{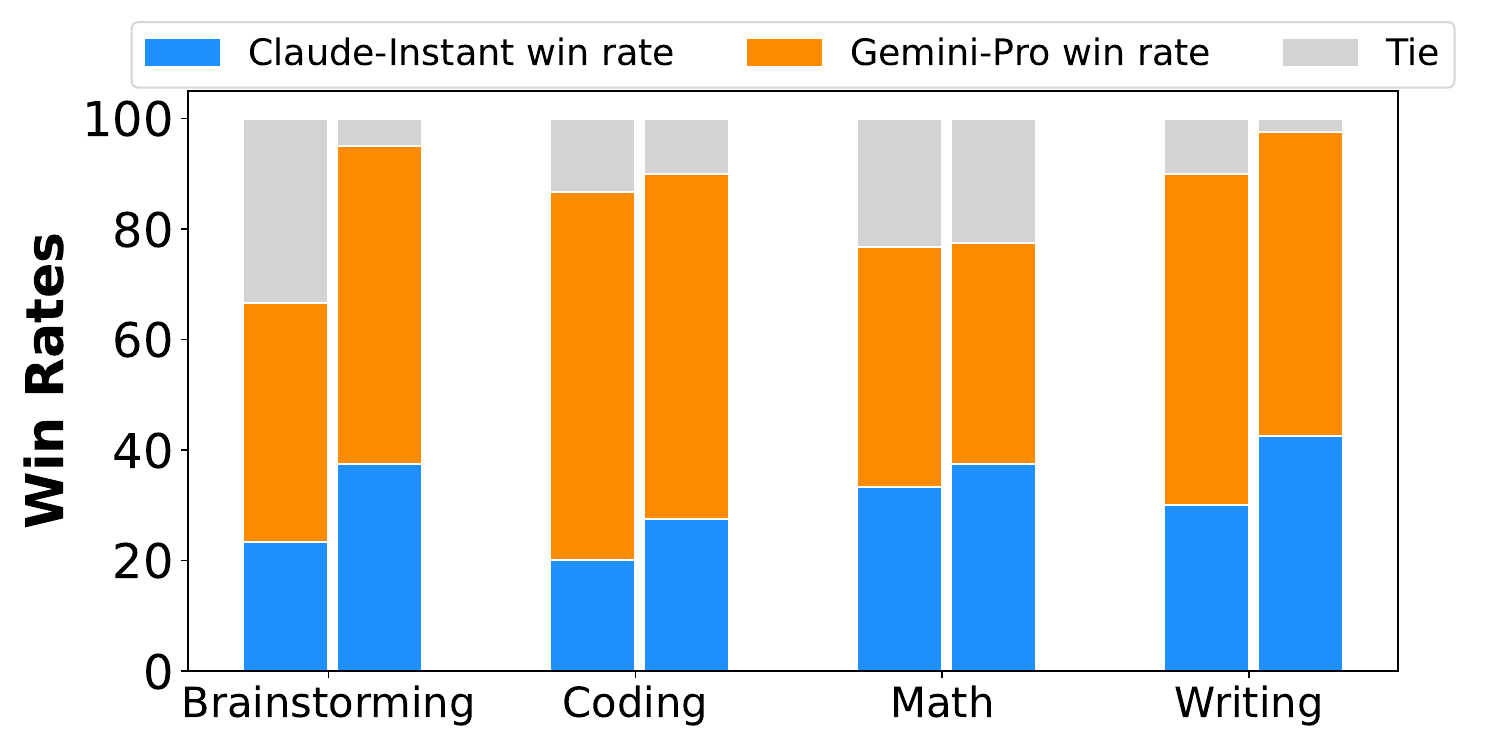}
        \caption{Claude-Instant vs. Gemini-Pro}
        \label{fig:winrates2}
    \end{subfigure}
    \hfill
    \begin{subfigure}[b]{0.31\textwidth}
        \includegraphics[width=\textwidth]{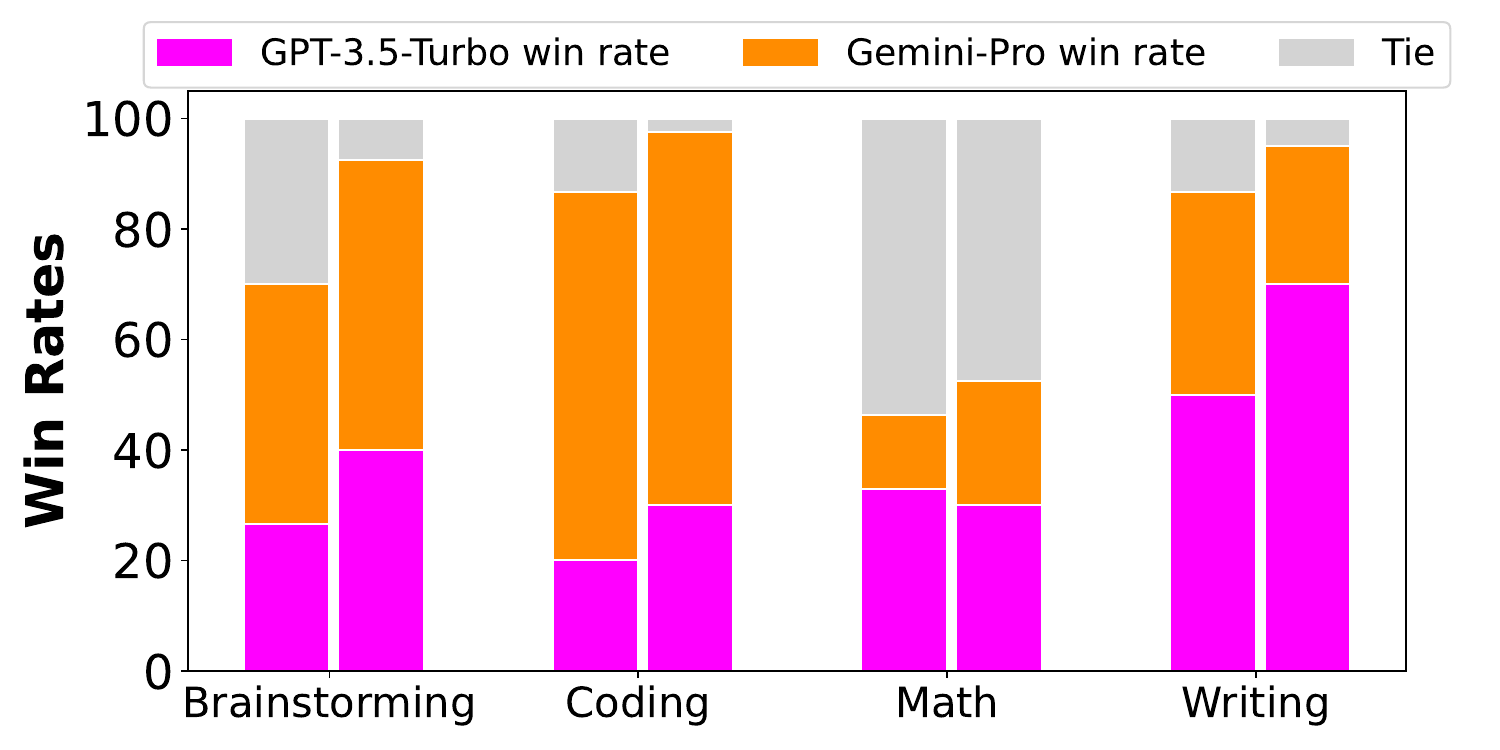}
        \caption{GPT-3.5-Turbo vs. Gemini-Pro}
        \label{fig:winrates3}
    \end{subfigure}
    \caption{System-level agreement -- win rates for each LLM pairwise comparison. Left bars in each scenario represent human expert results; right bars represent \modelname's meta-evaluation results.}
    \label{fig:winrates}
\end{figure*}

Based on Fig. \ref{fig:winrates}, we notice a consistent "preference in the same direction" between human experts and multi-agent debates across \textbf{all} LLM pairwise comparisons and scenarios. Notably, \textit{gpt-3.5-turbo} is favored (higher win rates) in \textit{brainstorming, math,} and \textit{writing} scenarios when compared with \textit{claude-instant}. Similarly, \textit{gemini-pro} is also preferred over \textit{claude-instant} in all scenarios. When comparing \textit{gpt-3.5-turbo} with \textit{gemini-pro}, a varied pattern in decision outcomes is observed: both human experts and multi-agent systems agree that \textit{gpt-3.5-turbo} outperforms \textit{gemini-pro} in scenarios involving \textit{math} and \textit{writing}. Conversely, \textit{gemini-pro} is deemed superior in \textit{brainstorming} and \textit{coding} scenarios. The high agreement of multi-agent preferences with human expert judgement results verifies the reliability of using multiple LLMs agents as meta-evaluators in various user-defined scenarios.

\begin{table*}[!htbp]
\centering
\footnotesize
\begin{tabular}{@{}llccccc@{}}
\toprule
\textbf{LLM Pairwise Comparisons} & \textbf{Criterion} & \textbf{Scenario} & \textbf{Meta-Evaluation} & \textbf{GPT-4-Turbo} & \textbf{Claude-2} & \textbf{GPT-3.5-Turbo} \\ 
\midrule
GPT-3.5-Turbo vs. Claude-Instant & Helpfulness & Brainstorming & 0.600 & \textbf{0.633} & 0.433 & 0.267 \\
 & Interpretability & Coding & \textbf{0.733} & 0.700 & 0.533 & 0.567 \\
 & Reasoning	& Math & \textbf{0.867} & 0.600 & 0.400 & 0.367\\
 & Creativity & Writing & \textbf{0.700} & 0.667 & 0.400 & 0.333\\ 
\midrule
Claude-Instant vs. Gemini-Pro & Helpfulness & Brainstorming & \textbf{0.667} & 0.533 & 0.467 & 0.500\\
 & Interpretability & Coding & \textbf{0.833} & 0.600 & 0.500 & 0.567\\
 & Reasoning	& Math & \textbf{0.767} & 0.667 & 0.330 & 0.367\\
 & Creativity & Writing & \textbf{0.733} & 0.633 & 0.400 & 0.500\\
\midrule
GPT-3.5-Turbo vs. Gemini-Pro & Helpfulness & Brainstorming & \textbf{0.733} & 0.600 & 0.467 & 0.467\\
 & Interpretability & Coding & \textbf{0.833} & 0.733 & 0.567 & 0.667\\
 & Reasoning	& Math & \textbf{0.867} & 0.767 & 0.500 & 0.433\\
 & Creativity & Writing & \textbf{0.767} & 0.667 & 0.500 & 0.433\\
\bottomrule
\end{tabular}
\caption{Example-level agreement rate comparison between human expert and \modelname's meta-evaluation vs. human expert and single LLM evaluation across four scenarios and criteria.}
\label{table:baseline exp}
\end{table*}

\begin{figure}[!htbp]
  \centering
  \includegraphics[width=\linewidth, height=3cm]{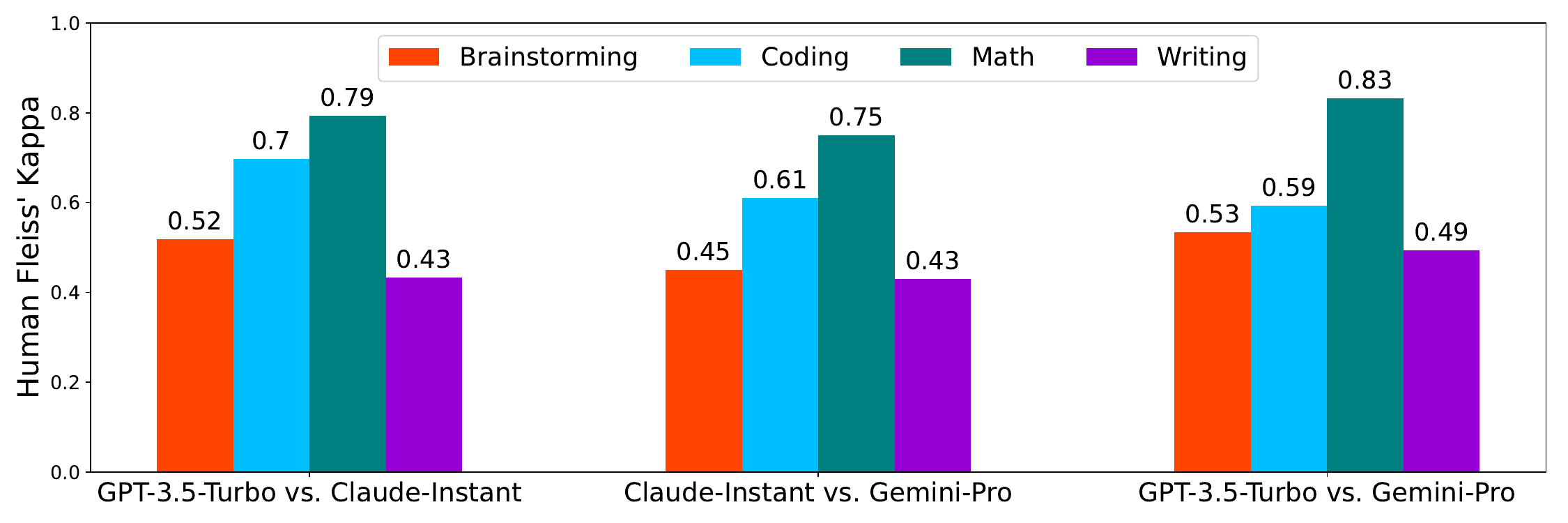}
  \caption{Human Fleiss Kappa for each LLM pairwise comparison under four scenarios.}
  \label{fig:fleiss.pdf}
\end{figure}

\section{Exp-II: Meta-Evaluation vs. LLM Evaluators}
\label{sec:metaeval}

Next, we use the fact that \modelname allows for reliable and scalable meta-evaluation to examine the traits of LLMs as evaluators.

\paragraph{Q2: \textbf{\textit{What are the capabilities and limitations of each LLM evaluator?}}} To effectively evaluate the performance of each LLM in its role as an evaluator, we adopt an approach that involves comparing the outcomes from our meta-evaluation process with the evaluations made independently by each LLM evaluator, which uncovers any disagreements or alignments between them. In the process, we aim to shed light on the performance characteristics of each LLM evaluator, which helps us identify which of them demonstrate superior evaluative abilities, thereby contributing to our understanding of their reliability in evaluating responses under each scenario. In addition, we provide a comprehensive cost-performance analysis to decide which LLM evaluator is the most suitable choice in each scenario.

\paragraph{Setup} For meta-evaluation, we employed three LLMs (\textit{gpt-4-turbo}, \textit{claude-2}, and \textit{gpt-3.5-turbo}) as evaluators to perform pairwise comparisons of responses from three distinct LLMs: \textit{gpt-3.5-turbo}, \textit{claude-instant}, and \textit{gemini-pro}. Previous studies have highlighted the presence of positional biases when LLMs are used as evaluators \cite{Wang2023LargeLM}. In response to these findings, we have implemented a strategy of randomization to mitigate such biases. Specifically, the sequence in which submissions from LLMs are presented to the agent evaluators is randomized. Additionally, we also randomize the order of discussions for each agent evaluator in every case. These approaches ensure that the process is fair and unbiased as much as possible, allowing for a more accurate assessment of the LLM evaluators' performance. The meta-evaluations were done under the following 8 scenarios: \textit{brainstorming, coding, dialog, judgement, open-domain general, open-domain science, and writing}, with the same set of 4 criteria used during human expert annotation.

\paragraph{Results} Table \ref{table:multiagent exp} compares the agreement rate between \modelname's meta-evaluation and each LLM evaluator across criteria and scenarios. We observe that \textit{gpt-4-turbo}, when serving as an evaluator, has the highest agreement rates with our meta-evaluation, particularly in the scenarios of \textit{brainstorming, dialog}, and \textit{ODG} with the \textit{helpfulness} criterion. It stands out with the highest overall average score of 0.780. However, our selected open-source model evaluator, \textit{auto-j}, outperforms \textit{gpt-4-turbo} in evaluating \textit{coding} questions based on the \textit{helpfulness} criterion. In addition, it exhibits the highest agreement rate with our meta-evaluation in the \textit{judgement} scenario, according to the \textit{helpfulness} criterion, indicating it as the most capable evaluator in this setting. It also achieves comparable results with other closed-source models like \textit{claude-2} and \textit{gpt-3.5-turbo} in most of the other scenarios. 

While \textit{gpt-4-turbo} performs the best as an evaluator in a majority of scenarios, it is not necessarily the best choice when we take into consideration its relatively high API costs. In fact, both the more affordable version (\textit{gpt-3.5-turbo}) and our selected free, open-source model (\textit{auto-j)} show comparable performance in scenarios like \textit{judgement} and \textit{writing}. For coding-related evaluations, the slightly less expensive \textit{claude-2} could be a more cost-effective alternative to \textit{gpt-4-turbo}.

\begin{table}[!htbp]
\centering
\tiny
\begin{tabular}{@{}llcccc@{}}
\toprule
\textbf{Criterion} & \textbf{Scenario} & \textbf{GPT-4-Turbo} & \textbf{Claude-2} & \textbf{GPT-3.5-Turbo} & \textbf{Auto-J}\\ 
\midrule
Helpfulness & Brainstorming & \textbf{0.800} & 0.500 & 0.650 & 0.575\\
 & Coding & 0.600 & \textbf{0.725} & 0.675 & 0.675\\
 & Dialog & \textbf{0.800} & 0.700 & 0.700 & 0.625\\
 & Judgement & 0.725 & 0.625 & 0.725 & \textbf{0.750}\\ 
 & Math & \textbf{0.825} & 0.650 & 0.600 & 0.350\\
 & ODG & \textbf{0.850} & 0.525 & 0.575 & 0.700\\
 & ODS & \textbf{0.875} & 0.525 & 0.575 & 0.675\\
 & Writing & \textbf{0.750} & 0.600 & \textbf{0.750} & 0.600\\ 
\midrule
Interpretability & Coding & \textbf{0.825} & 0.600 & 0.550 & 0.525\\
\midrule
Reasoning & Math & \textbf{0.650} & 0.525 & 0.475 & 0.450\\
 & Judgement & \textbf{0.750} & 0.650 & 0.700 & 0.675\\ 
\midrule
Creativity & Writing & \textbf{0.775} & 0.600 & 0.575 & 0.650\\
 & Brainstorming & \textbf{0.800} & 0.525 & 0.550 & 0.625\\
 & Dialog & \textbf{0.875} & 0.750 & 0.700 & 0.800\\
 \midrule
 Average & Overall & \textbf{0.780} & 0.607 & 0.629 & 0.619\\
\bottomrule
\end{tabular}
\caption{Agreement rate between \modelname's meta-evaluation and each LLM evaluator for comparing GPT3.5-Turbo vs. Claude-Instant.}
\label{table:multiagent exp}
\end{table}

\begin{table*}[t!]
\centering
\footnotesize
\begin{tabular}{@{}llcccc@{}}
\toprule
\textbf{Criteria Format} & \textbf{Criteria} & \textbf{Scenario} & \textbf{GPT-4-Turbo} & \textbf{Claude-2} & \textbf{GPT-3.5-Turbo} \\ 
\midrule
General & Helpfulness & Brainstorming & \textbf{0.800} & 0.500 & \textbf{0.650}\\
 & Interpretability & Coding & \textbf{0.825} & \textbf{0.600} & \textbf{0.550}\\
 & Reasoning	& Math & \textbf{0.650} & \textbf{0.525} & 0.475\\
 & Creativity & Writing & \textbf{0.800} & \textbf{0.600} & \textbf{0.575}\\
\midrule
Shortened & Helpfulness & Brainstorming & 0.675 & 0.500 & 0.575\\
 & Interpretability & Coding & 0.675 & 0.325 & 0.425 \\
 & Reasoning	& Math & 0.625 & 0.425 & 0.400\\
 & Creativity & Writing & 0.675 & 0.250 & 0.525\\
\midrule
Gibberish & Helpfulness & Brainstorming & 0.575 & 0.450 & 0.575 \\
 & Interpretability & Coding & 0.700 & 0.275 & 0.525 \\
 & Reasoning	& Math & \textbf{0.650} & 0.200 & 0.400 \\
 & Creativity & Writing & 0.550 & 0.150 & 0.450\\ 
\midrule
Shuffled & Helpfulness & Brainstorming & 0.625 & \textbf{0.550} & 0.500 \\
 & Interpretability & Coding & 0.600 & 0.400 & 0.525 \\
 & Reasoning	& Math & 0.625 & 0.225 & \textbf{0.600}\\
 & Creativity & Writing & 0.625 & 0.275 & 0.500\\
 \midrule
Flipped & Helpfulness & Brainstorming & 0.725 & 0.325 & 0.550 \\
 & Interpretability & Coding & 0.725 & 0.425 & 0.300\\
 & Reasoning	& Math & 0.575 & 0.250 & 0.500\\
 & Creativity & Writing & 0.750 & 0.075 & 0.550\\
 \midrule
 Masked & Helpfulness & Brainstorming & 0.725 & 0.300 & 0.500 \\
 & Interpretability & Coding & 0.650 & 0.225 & 0.475\\
 & Reasoning	& Math & 0.575 & 0.150 & 0.375\\
 & Creativity & Writing & 0.575 & 0.200 & 0.400\\
\bottomrule
\end{tabular}
\caption{Agreement rate between \modelname's meta-evaluation results and each LLM evaluator under various criteria prompt formats and scenarios comparing GPT3.5-Turbo vs. Claude-Instant.}
\label{criteria_variations}
\end{table*}

\section{Exp-III: Meta-Evaluation with Criteria Prompt Format Variations}
\label{sec:criteria_prompts}
\paragraph{Q3: \textbf{\textit{How do the qualities of criteria prompts influence the robustness of LLMs as evaluators in different scenarios?}}} Prior studies have revealed that variations in prompts can substantially affect the behavior of LLMs, particularly with the text they generate. With this in mind, we define various formatted criteria for evaluating LLM responses under each scenario. This approach aims to examine the extent to which different formats of criteria prompts influence both the performance and robustness of LLMs as evaluators.

\paragraph{Setup} We define five variations of the same criteria prompts: \textit{shortened, gibberish, shuffled, flipped,} and \textit{masked} (see Table \ref{tab:criteria-format} under Appendix \ref{sec:appendix:a} for detailed format). With these criteria format variations, we intend to observe how the LLMs as evaluators would respond differently when conducting evaluation. We compare the example-level agreement rate between \modelname's meta-evaluation results and each LLM evaluator.

\paragraph{Results}
Based on Table \ref{criteria_variations}, we observe that the performance of LLMs as evaluators generally deteriorates when certain letters in the criteria prompts are masked. Furthermore, the removal of guiding phrases at the beginning, such as "Not Helpful" or "Highly Helpful", can also diminish their effectiveness as evaluators. Both \textit{gpt-4-turbo} and \textit{gpt-3.5-turbo} demonstrate some resilience to these adversarially formatted criteria prompts, maintaining a relatively consistent agreement rates across various criteria formats. In contrast, \textit{Claude-2} often showcases confusion and refuses to evaluate, particularly in cases with gibberish and masked criteria prompts, where it rejects answering about half of the questions. It typically responds with statements like, \textit{"Unfortunately I do not have enough information here to provide a fair evaluation... The criteria describe different quality levels, but there is no detail on what specific aspects of the responses should be assessed... any judgement risks being arbitrary or biased..."}. None of the LLMs as evaluators we tested maintained very similar evaluation capabilities when faced with these adversarially formatted criteria prompts, indicating a limitation in these LLMs as evaluators' current design and application. Despite their advanced capabilities in fulfilling a variety of tasks, they may still struggle with understanding and responding accurately to substituted criteria information, highlighting an area for potential improvement in future iterations of LLM technology. Among all the different formatted criteria, we highlight the cases where the LLMs perform the best as evaluators in Table \ref{criteria_variations}.

\section{Conclusion}
In this work, we propose \modelname, a scalable, agent-debate assisted meta-evaluation framework for assessing the reliability and robustness of LLMs as evaluators. This approach addresses the expensive and time-intensive challenges inherent in traditional meta-evaluation methods, particularly pertinent as the usage of LLMs expands, necessitating a more scalable solution. Through our research, we have not only demonstrated the reliability of our proposed meta-evaluation framework, but also shed light on the capabilities and limitations of LLMs as evaluators in various scenarios. We observe how the results from these LLMs as evaluators vary based on modifications to the same criteria prompts. By open-sourcing our framework, we aim to foster further research in this field and encourage the development of more advanced and reliable LLMs as evaluators in the future. 


\newpage

\section*{Acknowledgements}
We thank  Chunting Zhou, Weizhe Yuan, Chunpu Xu, Yan Ma, and Binjie Wang for the helpful discussions and feedback.

\bibliography{anthology,custom}
\bibliographystyle{acl_natbib}

\clearpage
\appendix

\section{Meta-Evaluation Prompt}
\label{sec:appendix:a}

\begin{table*}[ht]
    \centering
    \small
    \begin{tabular}{>{\raggedright\arraybackslash\tt}p{0.98\textwidth}<{}}
        \toprule
        \color{brown}\textbf{<Initial Evaluation>} \\
            Compare the two submissions based on the criteria above.
            Which one is better? First, provide a step-by-step explanation of your evaluation
            reasoning according to the criteria. Avoid any potential bias. Ensure that
            the order in which the submissions were presented does not affect your judgement.
            Keep your explanation strictly under 150 words. Afterwards, choose one
            of the following options: \\
            Submission 1 is better: "1" \\
            Submission 2 is better: "2" \\
            Neither is better: "0" \\ \\
            
            Directly type in "1" or "2" or "0" (without quotes or punctuation) that corresponds to your reasoning. At the end, repeat just the number again by itself on a new line. \\
            
            [Question]: \{question\} \\
            
            [Submission 1]: \{submission\_1\} \\
            
            [Submission 2]: \{submission\_2\} \\
            
            [Criteria]: \{criteria\} \\
            
            [User]: \{user\_prompt\} \\ \\
        
            You are evaluating two submissions for a particular question,
            using a specific set of criteria. Above is the data.\\ \\

        \midrule
        \color{brown}\textbf{<Discussion Rounds>} \\
            \textbf{Always remember you are Speaker 1/2/3}. Review again your own previous evaluations/discussions first, then answer user's request from Speaker 1/2/3's perspective. \\

            [Question]: \{question\} \\
            
            [Submission 1]: \{submission\_1\} \\
            
            [Submission 2]: \{submission\_2\} \\
            
            [Criteria]: \{criteria\} \\
        
            [Speaker 1's Initial Evaluation]: \{evaluation\_1\} \\
        
            [Speaker 2's Initial Evaluation]: \{evaluation\_2\} \\
        
            [Speaker 3's Initial Evaluation]: \{evaluation\_3\} \\
        
            [Speaker \{speaker\_number\}'s Discussion -- Round \{round\_number\}]: \{discussion\_reasoning\} \\
            ...\\ \\
            
            Read the question, submissions, criteria, and evaluations above. First, explain your thoughts step-by-step about other speakers' evaluations. Second, explain your reasoning step-by-step regarding whether or not to change your original answer about which submission you think is better after considering other speakers' perspectives. Keep your reasoning strictly under 150 words. Afterwards, choose one of the following options: \\
            Submission 1 is better: "1" \\
            Submission 2 is better: "2" \\
            Neither is better: "0" \\ \\
        
            Directly type in "1" or "2" or "0" (without quotes or punctuation) that corresponds to your reasoning. At the end, repeat just the number again by itself on a new line. \\
            
        \bottomrule
    \end{tabular}
    \caption{Prompt template for meta-evaluation via multi-agent debate}
    \label{tab:metaeval-prompt}
\end{table*}

\begin{table*}[ht]
    \scriptsize
    \centering
    \begin{tabular}{>{\raggedright\arraybackslash\tt}p{0.98\textwidth}<{}}
        \toprule
        \color{brown}\textbf{<Type 1: General Format Version>} \\
             "1": "Not Helpful - The response is completely unrelated, lacks coherence, and fails to provide any meaningful information." \\
             "2": "Somewhat Helpful - The response bears some relevance but remains largely superficial and unclear, addressing only the peripheral aspects of the user's needs." \\
             "3": "Moderately Helpful - The response is mostly relevant and clear, covering the basic aspects of the query, but lacks depth and comprehensive elucidation." \\
             "4": "Helpful - The response is on-point, detailed, and well-articulated, offering valuable information and clarifications that meet the user's primary needs and enhance understanding." \\
             "5": "Highly Helpful - The response is exceptionally thorough and precise, providing additional insights and valuable supplementary information." \\ \\

        \midrule
        \color{brown}\textbf{<Type 2: Shortened Format Version>} \\
            "1": "The response is completely unrelated, lacks coherence, and fails to provide any meaningful information." \\
            "2": "The response bears some relevance but remains largely superficial and unclear, addressing only the peripheral aspects of the user's needs." \\
            "3": "The response is mostly relevant and clear, covering the basic aspects of the query, but lacks depth and comprehensive elucidation." \\
            "4": "The response is on-point, detailed, and well-articulated, offering valuable information and clarifications that meet the user's primary needs and enhance understanding." \\
            "5": "The response is exceptionally thorough and precise, providing additional insights and valuable supplementary information." \\ \\
            
        \midrule
        \color{brown}\textbf{<Type 3: Gibberish Format Version>} \\
            "1": "N*t H\$l\%ful - Th\$ r\$sp0n\$e is c mplt\$l? unr€la7\$d, la\$ks c()h\$r\$n(€, and f\#i/s t\# p\$o\&id\$ any m€an*\&gful !format\$on." \\
            "2": "S\#m\$*ha+ H\$\%*fu/ - Th\$ r\#s0!n\$ b\%ars \$o/e re\$ev*nc\$ b\$t r\$ma\$n\$ l\#rg\$l4 \$u/7\$r7cial an* !ncl=4r, a6r\$ss@n4 o7ly th\$ p\$r4ph@r\$l a5p\$cts \#f th\$ \$s*r's n**ds." \\
            "3": "M\$!7r\$t\#ly H\$lpfu\& - Th\$ r@s0*n\$@ !s m\$\%stl€ r\$'\$van7 an cl\$ar, c\$\%\$r\$n4 th\$ ba\$!c a\$\%cts of th\$ qu€ry, b\$t l\#cks d\$pth an cmpr\$h\$ns\$v\$ lu\$7\$dat!on." \\
            "4": "H\$lpfu\& - Th\$ r!s0*n\$e !s o/7-p\$!nt, d\$ta\$!l\$d, an w\$l/-a\&!u/at\$d, \#ff\$r!n4 v\#l\$\%bl\$ \#nformat\$on and cl*r\$!cat!ons th\#t m=t th\$ u/7\$r\'s pr!/ary n\$\$ds an* @n7anc\$ un\#rstand!n4." \\
            "5": "H4\#h7y H\$!p\%u\& - Th\$ r\$s\&*n!e !s \$xc\$pt\$\#nally th\#r\#7gh an* pr\$c\$\%\$, pr\#v\$d\$n\# a4*!t\$\#nal !ns\$4hts an* v\#lu\%bl\$ @*pp\%\$\%ntary \#n\%ormat\$on." \\ \\

        \midrule
        \color{brown}\textbf{<Type 4: Shuffled Format Version>} \\
            "1": "coherence fails provide unrelated, completely response - and the meaningful any to lacks Not Helpful is The information." \\
            "2": "superficial response largely addressing unclear, remains only needs. - relevance user's and the Helpful the peripheral some bears but aspects Somewhat The of" \\
            "3": "basic aspects query, lacks Moderately covering clear, - Helpful is depth response and comprehensive elucidation. relevant mostly the The and the of but" \\
            "4": "clarifications the is response information needs enhance and Helpful - on-point, valuable well-articulated, offering understanding. The and detailed, primary that user's meet" \\
            "5": "valuable Highly response is providing - the exceptionally Helpful information. insights thorough and additional precise, supplementary and The" \\ \\
            
        \midrule
        \color{brown}\textbf{<Type 5: Flipped Format Version>} \\
            "1": "toN lufpleH -  ehT esnopser si yletelpmoc detalernu, skcal ecnerehoc, dna sliaf ot edivorp yna lufgninaem noitamrofni." \\
            "2": "tamewoS lufpleH - ehT esnopser sraeb emos ecnaveler tub sniamer ylegral laicifrepus dna raelcnu, gnisserdda ylno eht larehpirep stcepsa fo eht s'resu sdeen." \\
            "3": "yletaredoM lufpleH - ehT esnopser si yltsom tnaveler dna raelc, gnirevoc eht cisab stcepsa fo eht yreuq, tub skcal htped dna evisneherpmoc noitadicule." \\
            "4": "lufpleH - ehT esnopser si tniop-no, deliated, dna detalucitra-llew, gnireffo elbaulav noitamrofni dna snoitacifralc taht teem eht s'resu yramirp sdeen dna ecnahne gnidnatsrednu." \\
            "5": "ylhgiH lufpleH - ehT esnopser si yllanoitpecxe hguoroht dna esicerp, gnidivorp lanoitidda sthgisni dna elbaulav yratnemelppus noitamrofni." \\ \\

        \midrule
        \color{brown}\textbf{<Type 6: Masked Format Version>} \\
            "1": "N\_\_ H\_l\_ful - The r\_\_pnse is c\_m\_\_et\_\_y unr\_l\_te\_, lacks \_ohe\_en\_e, \_nd \_ai\_s to p\_ov\_de \_ny m\_a\_\_ngfu\_ \_nfo\_ma\_ion." \\
            "2": "\_om\_w\_at He\_p\_ul - T\_e re\_ponse be\_rs \_ome rel\_\_a\_ce but r\_\_ains la\_\_ely s\_\_erfi\_\_al and u\_cle\_\_, ad\_res\_\_ng onl\_ \_he \_\_ri\_\_er\_l a\_pe\_ts of t\_\_ u\_e\_'s ne\_ds." \\
            "3": "Mod\_\_\_tely \_elp\_\_l - Th\_ \_esp\_\_se is mos\_\_y re\_\_va\_t an\_ \_le\_r, c\_v\_\_ing the ba\_ic \_spe\_ts of the q\_e\_y, but \_\_cks \_e\_th and co\_preh\_ns\_ve el\_c\_d\_t\_on." \\
            "4": "\_\_lpful - \_he respo\_se is on-p\_in\_, d\_\_\_iled, and we\_l-ar\_icu\_ated, of\_er\_ng val\_ab\_e \_\_for\_ation and cl\_r\_fi\_\_t\_ons t\_at mee\_ the \_se\_'s p\_im\_r\_ \_eeds and en\_\_nce u\_de\_\_tan\_ing." \\
            "5": "Hi\_h\_y H\_\_p\_ul - The r\_spon\_e is e\_c\_p\_io\_al\_\_ th\_r\_ugh and p\_ec\_se, pr\_vi\_ing a\_di\_\_on\_l ins\_g\_ts and va\_u\_b\_e \_upp\_e\_en\_a\_y inf\_rma\_io\_." \\ \\

        \bottomrule
    \end{tabular}
    \caption{Criteria prompt format variations for helpfulness}
    \label{tab:criteria-format}
\end{table*}

\end{document}